%% file: main.tex
\documentclass{article}

\usepackage{arxiv}
\usepackage{amsmath}      
\usepackage{amssymb}      
\usepackage{amsfonts}     
\usepackage{graphicx}     
\usepackage{booktabs}     
\usepackage{hyperref}     
\usepackage{url}          
\usepackage[numbers]{natbib}       
\usepackage{doi}          
\usepackage{algorithm2e,algorithmic}  
\usepackage{multirow}     
\usepackage{float}        
\usepackage{subfigure}    
\usepackage{multicol}     
\usepackage{enumitem}     
\usepackage{lastpage}     
\usepackage{xspace}       

\input{math_definition}   

\title{Federated Inverse Probability Treatment Weighting \\for Individual Treatment Effect Estimation}

\author{
    Changchang Yin \\
    Department of Computer Science and Engineering\\
    The Ohio State University\\
    Columbus, OH, USA\\
    \texttt{yin.731@osu.edu} \\
    \And
    Hong-You Chen \\
    Department of Computer Science and Engineering\\
    The Ohio State University\\
    Columbus, OH, USA\\
    \texttt{chen.9301@osu.edu} \\
    \And
    Wei-Lun Chao \\
    Department of Computer Science and Engineering\\
    The Ohio State University\\
    Columbus, OH, USA\\
    \texttt{chao.209@osu.edu} \\
    \And
    Ping Zhang\thanks{Corresponding Author e-mail: zhang.10631@osu.edu} \\
    Department of Computer Science and Engineering\\
    Department of Biomedical Informatics\\
    Translational Data Analytics Institute\\
    The Ohio State University\\
    Columbus, OH, USA\\
    \texttt{zhang.10631@osu.edu} \\
}

\begin{document}
\maketitle

\input{0_abstract}


\input{1_intro}

\input{8_framework}

\input{4_method}
\input{2_exp}
\input{7_related_work}
\input{6_conclusion}

\input{5_ethic}

\bibliographystyle{apalike}
\bibliography{references}

\section*{Acknowledgements}
This work was funded in part by the National Science Foundation under award number IIS-2145625, by the National Institute of Allergy and Infectious Diseases under award number R01AI188576, and by a Research Pilot Award from Translational Data Analytics Institute at The Ohio State University.

\section*{Author contributions}
PZ conceived the project. CY, HC, WC and PZ developed the method. CY conducted the experiments. CY, HC, WC and PZ analyzed the results. CY, HC, WC and PZ wrote the manuscript. All authors read and approved the final manuscript.

\section*{Competing interests}
The authors declare no competing interests. 

\section*{Additional information}
Correspondence and requests for materials should be addressed to PZ.

\newpage
\input{appendix}

\end{document}

%% file: math_definition.tex
\usepackage{amsmath,amsfonts}
\usepackage{amsopn}
\usepackage{bm} 
\usepackage{multirow}
\usepackage{tabularx}
\newcommand{\Ours}{\method{Fed-IPTW}\xspace}

\newcommand{\FedAvg}{\method{FedAvg}\xspace}

\newcommand{\vct}[1]{\boldsymbol{#1}} 

\newcommand{\ProbOpr}[1]{\mathbb{#1}}
\newcommand\independent{\protect\mathpalette{\protect\independenT}{\perp}}
\def\independenT#1#2{\mathrel{\rlap{$#1#2$}\mkern2mu{#1#2}}}

\newcommand{\expect}[2]{%
\ifthenelse{\equal{#2}{}}{\ProbOpr{E}_{#1}}
{\ifthenelse{\equal{#1}{}}{\ProbOpr{E}\left[#2\right]}{\ProbOpr{E}_{#1}\left[#2\right]}}} 

\DeclareMathOperator{\argmin}{arg\,min}

\newcommand{\vtheta}{\vct{\theta}}

\newcommand{\vx}{{\vct{x}}}

\newcommand{\vphi}{\vct{\phi}}

\newcommand{\sD}{\mathcal{D}}

\newcommand{\sL}{\mathcal{L}}

\newcommand{\eat}[1]{}
\newcommand{\method}[1]{\textsc{#1}}

\newcommand{\ie}{i.e.\xspace}
\newcommand{\eg}{e.g.\xspace}

\newtheorem{assumption}{Assumption}

%% file: 0_abstract.tex
\begin{abstract} 
Individual treatment effect (ITE) estimation is to evaluate the causal effects of treatment strategies  on some important outcomes, which is a crucial problem in healthcare. Most existing ITE estimation methods are designed for centralized settings. However, in real-world clinical scenarios, the raw data are usually not shareable among hospitals due to the potential privacy and security risks, which makes the methods not applicable. In this work, we study the ITE estimation task in a federated setting, which allows us to harness the decentralized data from multiple hospitals. Due to the unavoidable confounding bias in the collected data, a model directly learned from it would be inaccurate. One well-known solution is Inverse Probability Treatment Weighting (IPTW), which uses the conditional probability of treatment given the covariates to re-weight each training example. Applying IPTW in a federated setting, however, is non-trivial. We found that even with a well-estimated conditional probability, the local model training step using each hospital's data alone would still suffer from confounding bias. To address this, we propose \Ours, a novel algorithm to extend IPTW into a federated setting that enforces both global (over all the data) and local (within each hospital) decorrelation between covariates and treatments. We validated our approach on the task of comparing the treatment effects of mechanical ventilation on improving survival probability for patients with breadth difficulties in the intensive care unit (ICU). We conducted experiments on both  synthetic and real-world eICU datasets and the results show that \Ours outperform state-of-the-art methods on all the metrics 
on factual prediction and ITE estimation tasks, paving the way for personalized treatment strategy design in mechanical ventilation usage.

\end{abstract}

%% file: 1_intro.tex
\section{Introduction}
\label{s_intro}

Estimating the individual treatment effect (ITE) is crucial in healthcare~\cite{qian2021synctwin,shalit2017estimating}. It enables prescribing the right treatments (\eg, drugs, clinical procedures) to individuals based on their health statuses. One common approach to ITE estimation is randomized controlled trials (RCTs), which are conducted by randomly allocating patients (of similar statuses) to two groups, treating them differently (\ie, one group undergoes interventions and the other does not), and comparing them in terms of a measurable outcome (\eg, the survival or recovery rate). However, conducting RCTs is very expensive and time-consuming~\cite{yao2018representation,liu2020estimating}. A promising alternative is to estimate the ITE by learning from observational data --- the tuples collected from past patients who had recorded individual statuses (denoted as $X$), underwent treatments or not (denoted as $T=1$ or $T=0$), and outcomes (denoted as $Y$).

High-levelly, ITE estimation aims to learn a model to predict $E[Y|X, T=1] - E[Y|X, T=0]$. The challenge lies in the confounding bias in the collected data \cite{rosenbaum1983central,qian2021synctwin}. For instance, for severe patients whose survival chances are low, the treatment is usually given even though the majority of the outcomes might be negative. Directly learning from such data would lead to an inaccurate and biased estimation that the treatment is ineffective, even if it indeed increases the survival cases.
It is thus important to decorrelate the covariates and treatments while estimating ITE.
Inverse Probability Treatment Weighting (IPTW)~\cite{rosenbaum1983central} is a widely used technique for such a purpose, which first estimates the conditional probability of treatment given the covariates, denoted as $P(T|X)$, and then uses it to re-weight each training example $(x, t, y)$. 
 
In this paper, we aim to extend IPTW from its conventional use case, centralized learning, into a federated learning (FL) setting~\cite{li2020federated-survey,kairouz2019advances}. That is, we attempt to leverage patients' data collected from multiple hospitals to collaboratively train the ITE estimation model while keeping the data decentralized. Federated Averaging (\FedAvg)~\cite{FedAvg} is arguably the most popular FL algorithm, which learns a model iteratively --- between parallel local model training using each decentralized data source (often called a client) and global model aggregation at the server --- for multiple communication rounds. 

Suppose the conditional probability $P(T|X)$ is given, 
the standard application of \FedAvg to IPTW would involve using $P(T|X)$ to re-weight each example during local training of the ITE estimation model at each hospital. However, since each hospital is likely to have a different data distribution\footnote{This is a common challenge in FL~\cite{li2020federated-survey,kairouz2019advances}: as each client collects data individually, it is unlikely that their data distributions are identical. In ITE estimation, each hospital client $c$ may have a different strategy for assigning treatments. The conditional probability of each hospital, \ie $P(T|X, H=h_c)$,
thus can be different from each other and from the global $P(T|X)$.}, the covariates and treatments within each hospital may not be fully decorrelated by the global $P(T|X)$. As a result, the learned ITE estimation model would suffer from confounding bias during local training and become inaccurate.

  
To address this issue, we propose \Ours, a novel IPTW formulation for federated ITE estimation. 
During local training at hospital (client) $c$, \Ours replaces $P(T|X)$ with $P(T|X, H=h_c)$, the conditional probability at the hospital, for re-weighting ($H$ represents the treatment assignment strategy in hospitals). This ensures local decorrelation between covariates and treatments, preventing the local model from learning the confounding bias caused by treatment assignment strategy differences. 
To further ensure global decorrelation within data from all the hospitals so that the aggregated model is not biased, \Ours estimates a hospital-specific weight to adjust the importance of each local model during global aggregation. With this new formulation in local training and global aggregation, the learned ITE estimation model can be less biased and more accurate.

We investigate several ways to estimate the conditional probability of each hospital $P(T|X,H=h_c)$. Compared to estimating it independently at each hospital, which may suffer from scarce data, we found that estimating it collaboratively with all the hospitals can be more robust. This leads to an interesting combination of ``personalized'' federated learning~\cite{Kulkarni2020SurveyOP} and ``generic'' federated learning~\cite{chen2021bridging} to fulfill federated ITE estimation --- personalized FL for the conditional probability, followed by generic FL for the ITE estimation model.

We validate \Ours on both synthetic and real-world datasets (i.e., eICU~\cite{eicu}). With the two-level decorrelation, \Ours outperforms state-of-the-art methods in federated ITE estimation, demonstrating its effectiveness.
 
In sum, our contributions are three-folded:
\begin{itemize} 
    \item We identify the challenge of applying IPTW to ITE estimation in a federated setting. 
    \item We propose a new approach \Ours that promotes decorrelation between covariates and treatments both locally (within each hospital) and globally (over all the data). The resulting ITE estimation model thus suffers from less confounding bias and is more accurate.
    \item We conduct extensive experiments on both synthetic and real-world datasets to demonstrate the effectiveness of the proposed \Ours on the ITE estimation of mechanical ventilation on improving survival probability, paving the way for personalized treatment strategy design in mechanical ventilation usage.
\end{itemize}

%% file: 8_framework.tex
\begin{figure}[ht]
\centering
\includegraphics[width=0.7\linewidth]{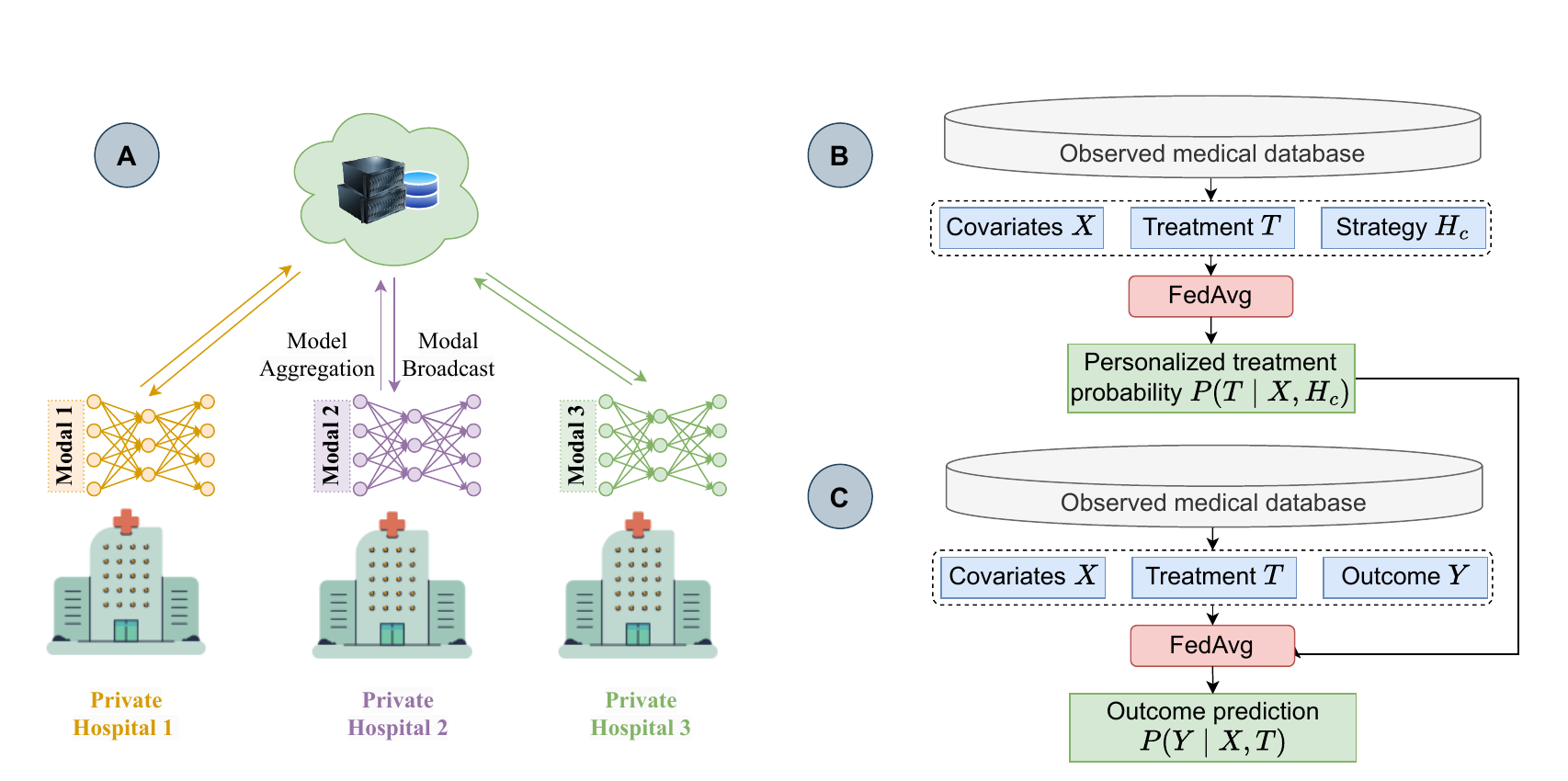}  

\caption{Framework of \Ours. (A) Federated learning in healthcare. (B) Treatment probability learning to remove local confounding bias. (C) Unbiased factual prediction learning for ITE estimation.} 
\label{fig:framework}
\end{figure} 

\section{Overall Framework}

We develop a novel framework \Ours for federated ITE estimation, as shown in \autoref{fig:framework}. 
Given the collected data distributed among $C$ clients (\eg, hospitals), each client $c\in[C]$ has a data set $\sD_c = \{(\vx_{c,i}, t_{c,i}, y_{c,i})\}_{i=1}^{|\sD_c|}$, where $\vx_{c,i}$, $t_{c,i}$, and $y_{c,i}$ denotes observed covariates, treatments and outcome for patient $i$. \Ours learns a model to estimate the individual treatment effect:
$E[Y|X=\vx_{c,i}, T=1] - E[Y|X=\vx_{c,i}, T=0]$.

\Ours is a federated learning framework, which solves optimization problems by iteratively repeating the local model training phase and global model aggregation phase, as shown in \autoref{fig:framework}(A).
In the local model training phase, the clients  train their local model in parallel using local private data.  
In the global model aggregation phase, the server will receive multiple local models from clients, aggregate them into a single global model and broadcast it to the clients to start the next round of local training.
The process will repeat many rounds until convergence. Specifically, in this study, we utilize \FedAvg \cite{FedAvg} to serve as the federated learning framework for ITE estimation.

We train \Ours with two-step \FedAvg.
In the first step, we adopt \FedAvg to learn patient-specific weight by estimating the treatment probability $P(T|X,H)$, as \autoref{fig:framework}(B) shows. 
During local training at hospital (client) $c$, \Ours learn a conditional probability $P(T|X, H=h_c)$ for each hospital $c$. $H$ represents the treatment assignment strategy in hospitals. This ensures local decorrelation between covariates and treatments, preventing the local model from learning the confounding bias caused by treatment assignment strategy differences.  

In the second step as \autoref{fig:framework}(C) shows, \Ours adopt \FedAvg to learn the unbiased factual prediction model $P(Y|X,T)$ by computing two weights for each sample (including a patient-specific weight and a hospital-specific weight). The patient-specific weight ensures the local decorrelation, while the hospital-specific weight that can adjust the importance of each local model during global aggregation can further ensure global decorrelation within data from all the hospitals so that the aggregated model is not biased. 
With the two weights, the learned ITE estimation model can be less biased in both local and global levels and become more accurate.

After the factual prediction model is well trained, we can easily estimate the ITE with $P(Y|X=x_{c,i}, T=1) - P(Y|X=x_{c,i}, T=0)$

%% file: 4_method.tex
\section{Method}

In this section, we first briefly review the ITE estimation task, IPTW and \FedAvg algorithms, and then present \Ours to estimate treatment effects in a federated setting robustly. 
We highlight the limitations (\ie, the existence of local confounding bias) inherent in the direct application of \FedAvg to IPTW.  Subsequently, we introduce the high-level idea of \Ours framework that decorrelates covariates and treatments at both local and global levels through the utilization of patient-specific and hospital-specific weights. Then we outline the methodology for calculating these two distinct weights.

\subsection{Background}
\label{s_background}

 
\subsubsection{Individual Treatment Effect (ITE) Estimation}

Given a patient $i$, the individual treatment effect estimation is defined as:
\begin{equation}
e_i = E[Y|X=\vx_i, T=1] - E[Y|X=\vx_i, T=0],
\end{equation}
where $X$ denotes the observed covariates (\eg, heart rate, respiratory rate), $T$ denotes a binary treatment variable (\eg, the usage of mechanical ventilator), $Y$ denotes a binary or continuous outcome (\eg, survival or length-of-stay in ICU). ITE estimation task aims to estimate how much outcome can be improved if the treatment is assigned to a specific patient.


In the collected clinical data, clinicians assign treatments based on the observed covariates, so $X$ and $T$ have strong correlations (\ie, confounding bias). 
Naively training a regression model to predict $E[Y|X, T]$ would cause significant estimation bias on the treatment effect.  



\subsubsection{IPTW in Centralized Setting}

IPTW is a widely used method to adjust the confounding bias when estimating ITE.
It re-weights each observed example based on the propensity score (\ie, the conditional probability of treatment given the covariates) as follows:
\begin{equation}
\label{eq:global_weight}
w_{i} = \frac{E[T] * I[t_i=1]}{P(T=1|\vx_i)} + \frac{(1 - E[T])*I[t_i=0]}{P(T=0|\vx_i)},
\end{equation} 
where $P(T|X)$ is usually modeled with a propensity score function which predicts the probability of treatment assignment given a specific patient. 
$I[\cdot]$ is the indicator function that returns 1 if the statement is true; otherwise, 0.

Re-weighting the samples could decorrelate the covariates $X$ and treatments $T$ by structuring the pseudo-population. We can directly train a factual prediction model on the pseudo-population to estimate the treatment effects:
\begin{equation}
\label{eq:tee_global}  
e_i = g_{\vphi} (X=\vx_i, T=1) - g_{\vphi}(X=\vx_i, T=0),
\end{equation} 
where $g_{\vphi}(X,T)$ denotes the factual prediction function that models the posterior outcome probability $P(Y|X,T)$.
The learnable parameter $\vphi$ is trained with weighted loss:
\begin{equation}
\label{eq:global_loss}
\ell(\vx_i, t_i, y_i, w_i; \vphi) = w_i * \ell_{bce}(g_{\vphi}(\vx_i, t_i), y_i). 
\end{equation} 
The weight $w_i$ is computed with \autoref{eq:global_weight}. $\ell_{bce}$ denotes binary cross entropy loss:
\begin{equation}
\label{eq:bce_loss}
\ell_{bce}(p, y) = -  y * \log (p) - (1 - y) * \log(1 - p) ,
\end{equation} 
where $p$ and $y$ denote the probability and ground truth sent to the loss function $\ell$. IPTW can easily handle continuous outcomes by replacing the binary cross entropy loss with mean squared error loss.


\subsubsection{Federated ITE Estimation}
Following \cite{hernan2010causal}, we assume the important assumptions (\ie, consistency, positivity, strong ignorability) are held in each client, which makes the ITE identifiable with IPTW~\cite{rosenbaum1983central}. Moreover, we have another assumption that the data among different clients have a noticable overlap, which makes the federated framework able to improve the model performance with collaboration among different clients.
The definition of the assumption can be found in \autoref{sec:assumptions} in supplementary materials.

In practical applications where the data are collected from privacy-sensitive users, the training data are distributed among $C$ clients (\eg, hospitals) that cannot be shared. Each client $c\in[C]$ has a training set $\sD_c = \{(\vx_{c,i}, t_{c,i}, y_{c,i})\}_{i=1}^{|\sD_c|}$. The goal of \textbf{federated learning (FL)} is to solve the following optimization problem, for harnessing the collective wisdom of clients to train a single model parameterized by $\vphi$;
\vspace{-2mm}
\begin{align}
& \min_{\vphi}~\sL(\vphi) = \sum_{c=1}^C \frac{|\sD_c|}{|\sD|} \sL_c(\vphi), \vspace{-4mm} 
\label{eq:iptw_fl}\\
& \text{where} \hspace{10pt} \sL_c(\vphi) = \frac{1}{|\sD_c|} \sum_{i=1}^{|\sD_c|} \ell(\vx_{c,i}, t_{c,i}, y_{c,i}, w^g_{c,i}; \vphi), \vspace{-2mm} \nonumber\\
& 
w^g_{c,i} = \frac{E[T] * I[t_{c,i}=1]}{P(T=1|\vx_{c,i})} + \frac{(1 - E[T]) * I[t_{c,i}=0]}{P(T=0|\vx_{c,i})} \vspace{-2mm} \nonumber
\end{align} 
Here, $|\sD| = \Sigma_c |\sD_c|$ is the training data size summed over all clients.  Based on the loss function on a sample $\ell$, $\sL$ and $\sL_c$ are the empirical risk on $\sD$ and $\sD_c$, respectively.   
Note that we introduce a weight $w^g_{c,i}$ to globally decorrelate covariates $X$ and treatments $T$ (\ie, among all hospitals). 
We use the superscript $g$ to emphasize that the global propensity score $P(T|X)$ is estimated for the whole patient population.


\paragraph{Federated averaging (\FedAvg).} 
Unfortunately, \autoref{eq:iptw_fl} cannot be solved directly since clients' data are separated and non-shareable. 
The standard FL algorithm to relax it is \FedAvg \cite{FedAvg}. Initialized from a model $\bar{\vphi}^{(0)}$, \FedAvg involves two steps repeated over multiple rounds of communication $r=1, ..., R$: parallel \emph{local training} at the clients to minimize each client's empirical risk (for several epochs) and \emph{global aggregation} at the server that takes an element-wise average over local model parameters, 
\vspace{-1mm}
\begin{equation}
    \label{eq:gfl_local}
\hspace{-2mm} \textbf{ Local:  } \tilde{\vphi}_c^{(r)} = \argmin_{\vphi} \sL_c(\vphi), \text{ initialized by } \bar{\vphi}^{(r-1)} 
\vspace{-4mm}
\vspace{-5pt}
\end{equation}
\begin{align}
\label{eq:gfl_global}
\hspace{-4mm} \textbf{ Global: } \bar{\vphi}^{(r)} \leftarrow \sum_{c=1}^C \frac{|\sD_c|}{|\sD|}{\tilde{\vphi}_c^{(r)}} \qquad \qquad \qquad \qquad \quad 
\vspace{-3mm}
\end{align}
That is, after each round, the server will receive multiple local models from clients, aggregate them into a single global model $\bar{\vphi}^{(r)}$, and broadcast it to the clients to start the next round of local training. The final global model $\bar{\vphi}^{(R)}$ will be used to estimate the ITE for the test or future patients.


\subsection{Limitation of IPTW with \FedAvg}

The formulation of \FedAvg in \autoref{eq:gfl_local} and \autoref{eq:gfl_global} strictly follows the IPTW objective defined in \autoref{eq:iptw_fl}. However, we argue that the local training step with each hospital's data would, unfortunately, suffer from confounding bias. 
More specifically, in the local training step, $w^g_{c,i}$ is used to re-weight data. While $w^g_{c,i}$ could guarantee global decorrelation, it may not decorrelate covariates and treatments within the data of each hospital. This is because different hospitals are unlikely to have the same treatment assignment strategy (\ie, the conditional probability of treatments given the covariates). 
For instance, in eICU \cite{eicu}, the probabilities of assigning the mechanical ventilation treatment to patients with potential breathing difficulties vary from 3\% to 33\% across hospitals, even though there is a noticeable overlap of the covariate distributions among hospitals. 
As a result, the global decorrelation does not guarantee the local decorrelation; the local confounding bias may still exist and would be learned during local training in \autoref{eq:gfl_local}. The bias could be propagated to the global model in the aggregation phase and make the estimated ITE inaccurate.
In this section, we propose a novel \Ours to achieve both global and local decorrelation when learning ITE estimation models.

\vspace{-2mm}
\subsection{Global and Local Correlation}
To better introduce the proposed \Ours, we first define the global correlation $COV(X, T)$ and local correlation $COV(X_c, T_c)$ as follows:
\vspace{-2mm}
\begin{align}
& COV(X,T) = \sum_{c=1}^C \sum_{i=1}^{|D_c|} w_{c,i} (\vx_{c,i} - \overline{X}) (t_{c,i} - \overline{T}),     \label{eq:global_correlation} \vspace{-2mm}\\
& \hspace{5pt} \text{where } \overline{X} = \frac{1}{w^s}\sum_{c=1}^C \sum_{i=1}^{|D_c|} w_{c,i} \vx_{c,i}, \quad  
        \overline{T} = \frac{1}{w^s}\sum_{c=1}^C \sum_{i=1}^{|D_c|} w_{c,i} t_{c,i}, \nonumber \vspace{-2mm}\\ 
& \hspace{30pt}  w^s = \sum_{c=1}^C w^s_c, \quad w^s_c = \sum_{i=1}^{|D_c|} w_{c,i}. \nonumber\\ 
& COV(X_c,T_c) = \sum_{i=1}^{|D_c|} w_{c,i} (\vx_{c,i} - \overline{X_c}) (t_{c,i} - \overline{T_c}), \label{eq:local_correlation} \vspace{-2mm}\\
& \hspace{5pt} \text{where } \overline{X_c} =  \frac{1}{w^s_c}\sum_{i=1}^{|D_c|} w_{c,i} \vx_{c,i}, \quad 
        \overline{T_c} = \frac{1}{w^s_c}\sum_{i=1}^{|D_c|} w_{c,i} t_{c,i}. \nonumber \vspace{-5mm}
\end{align}
$\overline{X_c}$ and $\overline{T_c}$ are the average covariates and treatments in client $c$. 
$\overline{X}$ and $\overline{T}$ are the average covariates and treatments among all the clients. 

To distinguish from $w^g_{c, i}$, which cannot achieve local decorrelation, we introduce new weights to achieve both global (\ie, $COV(X,T)=0$)  and local (\ie, $COV(X_c, T_c)=0)$) decorrelation in following subsections.
Note that due to the space limitation, we put the theoretical proof of two-level decorrelation in supplementary materials.

\subsection{Local Decorrelation} 
To remove the confounding bias during local training in \autoref{eq:gfl_local}, inspired by \cite{cadene2019rubi},
we introduce a new variable $h_c$ to capture local treatment strategies, so the FL algorithm can be encouraged to learn a generalized knowledge $P(T|X,H=h_c)$ (without overfitting local treatment assignment strategy), leading to more accurate estimation of conditional probability for the computation of weight $w_{c,i}$:
\begin{equation}
\label{eq:local_weight}
\hspace{-2mm} w_{c,i} = \frac{E[T_c] * I[t_i=1]}{ P(T=1|\vx_{c,i},h_c)} + \frac{(1 - E[T_c]) * I[t_i=0]}{P(T=0|\vx_{c,i},h_c) },
\end{equation} 
where, $X_c$ and $T_c$ are the covariates and treatments in client $c$. 
Different from \autoref{eq:iptw_fl}, we use $P(T|X,H=h_c)$ (referred as $P(T|X,h_c)$ for brevity) rather than $P(T|X)$. 
$P(T|X, h_c)$ fits the posterior treatment distribution in local client $c$. For brevity, we leave the estimation of $P(T|X, h_c)$ in the later subsections. Note that in our implementation, $h_c$ is a scalar variable and will be learned during the model training.


With the new patient-specific weight $w_{c, i}$ in \autoref{eq:local_weight}, the local correlation in \autoref{eq:local_correlation} can be re-written as:
\vspace{-2mm}
\begin{equation}
\label{eq:local_correlation_result}
\begin{aligned}
 COV(X_c, T_c) = \sum_{i=1}^{|D_c|} (\vx_{c,i} 
- E[X_c]) (E[T_c] - E[T_c])=0.
\end{aligned}
\end{equation}  
That is, with the consideration of hospital treatment strategy, we decorrelate $X_c$ and $T_c$ locally. 
The detailed proof for \autoref{eq:local_correlation_result} can be found in \autoref{eq:app:local_decorrelation} the supplementary material.


With this new $w_{c, i}$, the factual prediction parameter $\vphi$ is updated locally with the following loss function:
\vspace{-1mm}
\begin{equation}
    \label{eq:phi_local_update}
    \hspace{-4mm} \ell(\vx_{c,i}, t_{c,i}, y_{c,i}) = w_{c,i} \ell_{bce}(g_{\vphi}(\vx_{c,i}, t_{c,i}), y_{c,i}) 
    \vspace{-2mm}
\end{equation}

\begin{figure} 
\vspace{-4mm}
\centering
\includegraphics[width=0.5\linewidth]{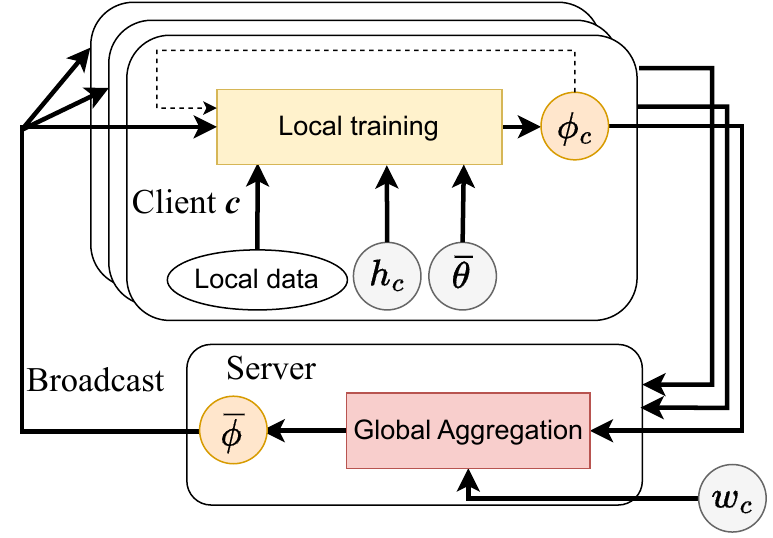}  
\vspace{-3mm}
\caption{Pipeline of factual prediction model training. The variables in gray color (\ie, $h_c$, $\overline{\vtheta}$ and $w_c$) are frozen during the training. $h_c$ and $\overline{\vtheta}$ is use to compute weight $w_{c,i}$.} 
\label{fig:factual_prediction_training}
\vspace{-4mm}
\end{figure}

\subsection{Global Decorrelation}


Re-weighting each patient's data with the new patient-specific weight $w_{c,i}$ can ensure local decorrelation. However, the global correlation in \autoref{eq:global_correlation} is not ensured to be 0.
Specifically, with $w_{c,i}$ defined in \autoref{eq:local_weight}, the global correlation between covariates $X$ and treatments $T$ becomes
\begin{equation}
\label{eq:global_correlation_wo_wc}
\begin{aligned}
\hspace{-5mm} COV(X, T)
= \sum_c |D_c|  (E[X_c] - E[X])(E[T_c] - E[T])
\end{aligned}
\end{equation}


To further decorrelate  $X$ and  $T$ globally (\ie, over all the hospitals), we compute a hospital-specific weight $w_c$:
\vspace{-2mm}
\begin{align}
\label{eq:hospital_weight}
w_c = \frac{P(T' = E[T_c])}{P(T'=E[T_c]|X'=E[X_c])},
\vspace{-2mm}
\end{align}
where $X'$ and $T'$ denote the average covariates and treatments in a hospital.
By augmenting this hospital-specific weight $w_c$ to the patient-specific weight $w_{c,i}$ defined in \autoref{eq:local_weight} --- namely, using $w_c \times w_{c,i}$ to re-weight each training data --- the global correlation $COV(X,T)$ in \autoref{eq:global_correlation} becomes
\vspace{-1mm}
\begin{align}
\label{eq:global_correlation_0}
\sum_c w_c |D_c|  (E[X_c] - E[X])(E[T_c] - E[T]) = 0.
\end{align}
The detailed proof can be found in \autoref{eq:app:global_decorrelation} in supplementary materials. Importantly, this combined weight ensures local decorrelation because $w_c$ is constant over all patients of the same hospital.
We leave the estimation of $w_c$ in the next subsections.




\textbf{Global aggregation.} 
In the implementation, we decompose the usage of $w_c$ and $w_{c,i}$ for training stability. We use $w_{c,i}$ alone to re-weight training data in local training and leave the use of 
$w_c$ in global aggregation. 
Namely, after the server receives local models $\tilde{\vphi}_c^{(r)}$ from clients, it aggregates them into a single global model via
\vspace{-1mm}
\begin{align}
 \bar{\vphi}^{(r)} \leftarrow \sum_{c=1}^C \frac{w_{c}|D_c|}{w_s}{\tilde{\vphi}_c^{(r)}} \text{, where }   w_{s} = \sum_{c=1}^C w_c |D_c|,
\label{eq_fed_agg}
\vspace{-2mm}
\end{align}
where we use $w_c$ to re-weight each local model.

To summarize, we show in \autoref{fig:factual_prediction_training} the training pipeline of the factual prediction model $g_{\vphi}$, which contains both the local training and global aggregation steps.

\begin{figure}
\centering
\vspace{-2mm}
\includegraphics[width=0.5\linewidth]{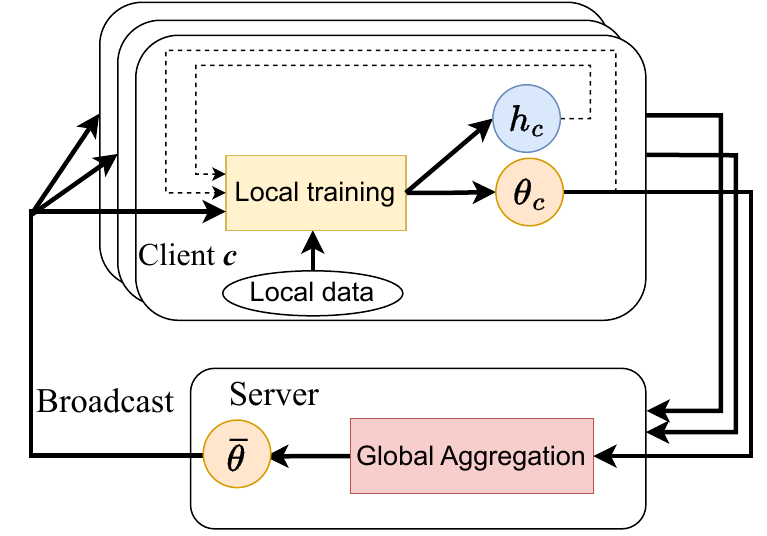}  
\vspace{-2mm}
\caption{Propensity score prediction model training pipeline. } 
\label{fig:propensity_score_training_pipeline}
\vspace{-3mm}
\end{figure}

\subsection{Patient-Specific Weight Learning} 

In this subsection, we show how to obtain faithful weights for \autoref{eq:local_weight} by learning a propensity score prediction function to model $P(T|X, H)$. 
Naively, one may learn a separate function for each hospital. However, due to the potentially limited sample size in each client, a separately trained function may overfit the dataset before capturing the statistical regularities between covariates and treatment.
Thus, we also adopt \FedAvg to train the functions with multiple hospitals collaboratively. Inspired by \cite{cadene2019rubi,chen2021bridging}, we model $P(T|X, H)$ by a function $f_{\vtheta}$ that takes both the covariates $X$ and the hospital variable $H$ as input. This hospital variable is expected to capture the variation of the treatment assignment strategies across hospitals, hence facilitating the global aggregation of 
the model parameters $\vtheta$.
It is worth noting that this FL process differs from and precedes the one for learning the factual prediction model. The learned $f_{\vtheta}$ can produce propensity scores 
differently for each hospital, reminiscent of personalized federated learning that learns for each client a personalized model.


\begin{algorithm}[t] 
\caption{\Ours}
\label{alg:fed_iptw}
\begin{algorithmic}[1]  
\STATE \textit{\# Treatment prediction learning with personalized FL}
\FOR{$r = 1, ..., R$}
\STATE Broadcast global model $\bar{\vtheta}^{(r-1)}$ to clients;
\FOR{$c = 1, ..., C$} 
\STATE Update treatment strategy $h_c$ with \autoref{eq:local_bias};
\STATE Train the propensity score prediction model $\tilde{\vtheta}_c^{(r)}$ with \autoref{eq:theta_update};
\ENDFOR
\STATE Aggregate $\tilde{\vtheta}_c^{(r)}$ into  $\bar{\vtheta}^{(r)}$ with \autoref{eq:theta_aggregation};
\ENDFOR 
\STATE \textit{\# Weight calculation}
\STATE Compute patient-specific weight $w_{c,i}$ with \autoref{eq:local_weight_f};
\STATE Compute hospital-specific weight $w_c$ with \autoref{eq:hospital_weight};
\STATE \textit{\# Factual prediction learning with generic FL}
\FOR{$r = 1, ..., R$} 
\STATE Broadcast global model $\bar{\vphi}^{(r-1)}$ to clients;
\FOR{$c = 1, ..., C$} 
\STATE Train local model $\tilde{\vphi}_c^{(r)}$  with \autoref{eq:phi_local_update};
\ENDFOR
\STATE Aggregate $\tilde{\vphi}_c^{(r)}$ into $\bar{\vphi}^{(r)}$ with \autoref{eq_fed_agg};
\ENDFOR

\end{algorithmic}
\end{algorithm}

\textbf{Local training.}
We use a binary cross-entropy loss to train the propensity score function:
\vspace{-1mm}
\begin{align}
\label{eq:fediptw_treatment_loss}
\sL_t(\vtheta, h_c) = \sum_{i=1}^{|D_c|} \ell_{bce}(f_{\vtheta}(\vx_{c,i}, h_c), t_{c,i}),  
\vspace{-3mm}
\end{align}
where $h_c$ denotes the \emph{learnable} parameter for hospital $c$ that captures the specific treatment strategy. 

We take a coordinate descent approach to learn $\vtheta$ and $h_c$. In each round of local training, given the initialized global model $\bar{\vtheta}^{(r-1)}$, we first freeze it and update $h_c$ alone.
\vspace{-1mm}
\begin{equation} 
\label{eq:local_bias}
h_c = \arg \min_{h_c} \sL_t(\bar{\vtheta}^{(r-1)}, h_c)
\vspace{-1mm}
\end{equation}


Then we freeze the learned $h_c$ and update the model weights $\vtheta$ to learn the shareable statistical regularities between covariates and treatments:
\begin{equation}
    \label{eq:theta_update}
    \tilde{\vtheta}_c^{(r)} = \argmin_{\vtheta} \sL_t(\vtheta, h_c), \text{ initialized by } \bar{\vtheta}^{(r-1)}
\end{equation}


\textbf{Global aggregation.} 
After each round of local training, the server receives multiple local models $\tilde{\vtheta}_c^{(r)}$ from clients and aggregates them into a global model $\overline{\vtheta}^{(r)}$.
\vspace{-1mm}
\begin{align}
\label{eq:theta_aggregation}
 \bar{\vtheta}^{(r)} \leftarrow \sum_{c=1}^C \frac{|\sD_c|}{|\sD|}{\tilde{\vtheta}_c^{(r)}}.
\vspace{-2mm}
\end{align}
Note that $h_c$ is kept in the local client without being shared with other clients or the server. 
\autoref{fig:propensity_score_training_pipeline} display the training pipeline of the propensity score prediction model $f_{\vtheta}$.

\textbf{Patient-specific weight computation.}
We compute the weight for each patient's data as follows:
\begin{equation}
\label{eq:local_weight_f}
\hspace{-5mm}w_{c,i} = \frac{E[T_c] * I[t_{c,i}=1]}{f_{\vtheta}(x_{c,i}, h_c)} + \frac{(1 - E[T_c])*I[t_{c,i}=0]}{1 - f_{\vtheta}(x_{c,i}, h_c)}
\vspace{-1mm}
\end{equation}

\subsection{Hospital-Specific Weight Learning}
The global factual prediction model $g_{\vphi}$ is aggregated based on hospital-specific weight $w_c$ in \autoref{eq:hospital_weight}.
Calculating $w_c$ requires the estimation of $P(T')$ and $P(T'|X')$, where $X'$ and $T'$ denote the averages of covariates and treatments, respectively, in a hospital. 
To this end, we follow the practice of existing federated ITE estimation work \cite{AdaTRANS,FedCI,causalRFF}, which allows clients to share the statistics information $(|D_c|, E[X_c], E[T_c])$ with the server.
By gathering this information, we can then use Gaussian distribution to characterize the distribution of the treatment average (\ie, $P(T')$) and utilize the Gaussian process to represent the posterior distribution of the treatment average conditioned on the covariate average (\ie, $P(T'|X')$). 
Please refer to the \autoref{sec:hospital_weight} in the supplementary material for details.


\vspace{-2mm}
\subsection{Overall \Ours Algorithm}
Algorithm \ref{alg:fed_iptw} describes the training process of propensity score prediction model $f_{\vtheta}$ and factual prediction model $g_{\vphi}$.

\vspace{-3mm}

%% file: 2_exp.tex
\section{Results}

In this section, we introduce the dataset we use for this study and then  demonstrate the effectiveness of \Ours in ITE estimation tasks.

\subsection{Datasets}
\textbf{Synthetic Dataset.}  
Obtaining ground truth for evaluating causal inference algorithms is a challenging task. Thus, most of the state-of-the-art methods are evaluated using synthetic or semi-synthetic datasets. Following \cite{causalRFF,AdaTRANS}, the synthetic data is simulated with the following distributions:
\begin{equation*}
\begin{aligned}
    z_{c,i} \sim Cat(\rho), \quad x_{c,ij} \sim Bern(\phi(a_{j0}+(z_{c,i})^\top a_{j1})), \\ t_{c,i} \sim Bern(\phi(b_0 + (z_{c,i})^\top (b_1 + \Delta))), \\
    y_{c,i}(0) \sim N(sp(c_0 + (z_{c,i})^\top (c_1 + \Delta)), \sigma_0^2), \\
    y_{c,i}(1) \sim N(sp(d_0 + (z_{c,i})^\top (d_1 + \Delta)), \sigma_1^2),
\end{aligned}
\end{equation*}
where $Cat(\cdot)$, $N(\cdot)$, and $Bern(\cdot)$ denote the categorical distribution, normal distribution, and Bernoulli
distribution, respectively. 
$\phi(\cdot)$ denotes the sigmoid function, $sp(\cdot)$ denotes the softplus function, and $x_{c,i} = [x_{c,i1},..., x_{c,id_x}]^\top \in R^{d_x}$ with $d_x=30$. 
For each source, we simulate 10 replications with $|D_c| = 1000$ records. We only keep  $(x_{c,i}, t_{c,i}, y_{c,i})$ as the observed data, where $y_{c,i}=y_{c,i}(0)$ if $t_{c,i}=0$ and $y_{c,i}=y_{c,i}(1)$ if $t_{c,i}=1$. 
To simulate data, we randomly set the ground truth parameters as follows: $\rho$ = [0.11, 0.22, 0.35, 0.25, 0.15],
($c_0$, $d_0$) = (0.85, 5.2), ($c_1$, $d_1$) are drawn i.i.d from $N(0, 2I)$, $a_{j0}$ and elements of $a_{j1}$ are drawn
i.i.d from $N(0, 2)$.

\textbf{eICU Datasets.} The eICU Collaborative Research Database~\cite{eicu} is a freely available multi-center database for critical care research with high granularity data for over 200,000 admissions to ICUs monitored by eICU Programs across the United States. 
In this work, we study the treatment effect of mechanical ventilation on the sepsis cohort. We use oxygen therapy (including supplemental oxygen and non-invasive ventilation as control treatments for mechanical ventilation treatments.
Following \cite{peine2021development,yin2022deconfounding}, we collect demographics (i.e., age, gender, race, height, weight) and 48 hours of EHR data (\ie, glasgow coma scale, heart rate, systolic, mean and diastolic blood pressure, respiratory rate, SpO2, temperature, potassium, sodium, chloride, glucose, BUN, creatinine, magnesium, calcium, carbon dioxide, SGOT, SGPT, total bilirubin, albumin, white blood cells count, PaO2, PaCO2, base excess,
bicarbonate, lactate, PaO2/FiO2 ratio) as covariates for each patient and use mechanical ventilation as treatment.  
For each vital sign and lab test, we use the last observed values during the first 48 hours as the observed covariates. If the patient received treatment in the first 48 hours, we only use the records before the mechanical ventilation usage.
We consider survival (or mortality) as a binary outcome.
All the patients with sepsis are extracted from eICU. We use the hospitals with more than 200 patients as clients. \autoref{tab:statis} in \autoref{sec:eicu_data} displays the statistics of the extracted dataset.

\textbf{Semi-synthetic Dataset based on eICU}.
We construct a semi-synthetic dataset based on a real-world dataset: eICU \cite{eicu}. We learn the distribution from eICU and generate data with the same 
simulation process to the way we generate the fully-synthetic dataset, with the exception that we only need to synthesize the potential outcomes.

 
\subsection{Compared Methods}
We compared the proposed method with state-of-the-art federated treatment effect estimation methods: AdaTrans~\cite{AdaTRANS}, FedCI~\cite{FedCI} and CausalRFF~\cite{causalRFF}.

To demonstrate the effectiveness of the proposed \Ours, we develop various versions of the proposed method.
\Ours~is the main version. 
IPTW$^l$ learns the posterior probability $P(T|X)$ locally and can only achieve local decorrelation between $T$ and $X$.
IPTW$^g$ learns the posterior probability $P(T|X)$ globally with FedAvg, and can only achieve global decorrelation.
\Ours~$^{-h}$ acheive both local and global decorrelation with two weight $w_{c,i}$ and $w_c$. The difference from \Ours is that \Ours$^{-h}$ does not take the hospital variable $H$ as input.  

Moreover, we compare the federated ITE estimation methods with two centralized ITE estimation models with IPTW: Global and Global$^{-h}$. Both Global and Global$^{-h}$ have access to data in all the hospitals. The difference is that Global  takes the hospital variable $H$ as input to capture the local treatment assignment strategies while Global$^{-h}$ does not.

\subsection{Evaluation Metrics} 
\textbf{Factual prediction evaluation.}
For the real-world dataset, we first evaluate the performance of factual prediction (\ie, treatment prediction and risk prediction) with area under the receiver operating characteristic curve  (AUROC) and area Under the precision-recall curve (AUPRC).


\noindent
\textbf{ITE estimation evaluation.}
To evaluate the estimated ITE, we use 3 metrics:
influence function-based precision of estimating heterogeneous effects (IF-PEHE), 
rooted Precision in Estimation of Heterogeneous Effect  ($\sqrt{\text{PEHE}}$) and mean absolute error (MAE) between ground truth and average treatment effect (ATE). The calculation details of the metrics can be found in \autoref{sec:metrics} in supplemental materials.

\begin{table*} [t]
\centering
\caption{Factual prediction and ITE estimation results.}  
\label{tab:result_real_ite} 
\begin{tabular}{l|ccc|cc|cc}
\toprule 
& \multicolumn{3}{c|}{eICU} &  \multicolumn{2}{c|}{Semi-synthetic} &  \multicolumn{2}{c}{Synthetic}\\
Method  & AUROC $\uparrow$ & AUPRC $\uparrow$& IF-PEHE $\downarrow$& $\sqrt{\text{PEHE}}$ $\downarrow$& MAE $\downarrow$& $\sqrt{\text{PEHE}}$ $\downarrow$& MAE  $\downarrow$ \\
\midrule
AdaTrans~\cite{AdaTRANS}  & .798 $\pm$ .010 & .568  $\pm$ .011 & .261   $\pm$ .013  & .596  $\pm$ .011 & .527  $\pm$ .010 & 3.46  $\pm$ .025 & 2.23  $\pm$ .028 \\
FedCI~\cite{FedCI} &  .812 $\pm$ .011 & .579  $\pm$ .010 & .245   $\pm$ .012 &.570  $\pm$ .010 & .503  $\pm$ .011 & 3.24  $\pm$ .024  & 2.10  $\pm$ .029 \\
CausalRFF~\cite{causalRFF} &  .818 $\pm$ .010 & .583  $\pm$ .009 & .231   $\pm$ .011 &  .558  $\pm$ .011 & .470  $\pm$ .011 & 2.20  $\pm$ .023 & 1.83  $\pm$ .022\\
\midrule
\Ours &  \textbf{.826 $\pm$ .009} & \textbf{.602  $\pm$ .008} & \textbf{.213   $\pm$ .011} &  \textbf{.532  $\pm$ .010} & \textbf{.447  $\pm$ .010} & \textbf{ 2.02  $\pm$ .021} & \textbf{ 1.42  $\pm$ .020} \\
\midrule
Global$^{-h}$ & .862 $\pm$ .010 & .655  $\pm$ .009 & .198   $\pm$ .013 &  .481  $\pm$ .011 & .394  $\pm$ .009 & 1.72  $\pm$ .022  & 1.31  $\pm$ .023 \\
Global & .881 $\pm$ .011 & .675  $\pm$ .009 & .182   $\pm$ .012 &  .475  $\pm$ .012 & .374  $\pm$ .010 & 1.53  $\pm$ .023  & 1.12  $\pm$ .025 \\
  \bottomrule
\end{tabular}  
\end{table*}

\begin{table*} [ht]
\centering
\caption{Ablation study.  GD and LD denote global decorrelation and local decorrelation respectively. }  
\label{tab:result_real_ablation} 
\setlength{\tabcolsep}{2.8pt}
\begin{tabular}{l|cc|ccc|cc|cc}
\toprule 
& \multicolumn{2}{c|}{Setting}  & \multicolumn{3}{c|}{eICU} &  \multicolumn{2}{c|}{Semi-synthetic} &  \multicolumn{2}{c}{Synthetic} \\
Method & GD & LD   &  AUROC $\uparrow$& AUPRC $\uparrow$& IF-PEHE $\downarrow$& $\sqrt{\text{PEHE}}$ $\downarrow$& MAE $\downarrow$& $\sqrt{\text{PEHE}}$ $\downarrow$& MAE $\downarrow$  \\
\midrule 
IPTW$^{l}$ & & \checkmark & .806 $\pm$ .009 & .594 $\pm$ .008 & .246 $\pm$ .012 &  .683  $\pm$ .010 & .592$\pm$ .011 & 3.32 $\pm$ .034 & 2.63$\pm$ .039\\ 
IPTW$^{g}$ & \checkmark & & .812 $\pm$ .011 & .593 $\pm$ .009 & .267 $\pm$ .011 & .642  $\pm$ .011 & .573 $\pm$ .011& 3.21 $\pm$ .028 & 2.52 $\pm$ .031 \\
\Ours$^{-h}$ &  \checkmark &\checkmark & .814 $\pm$ .009 & .595 $\pm$ .008 & .226 $\pm$ .011 & .562  $\pm$ .012 & .489 $\pm$ .010& 2.15 $\pm$ .023 & 1.63 $\pm$ .023 \\ 
\Ours & \checkmark & \checkmark &  \textbf{.826 $\pm$ .009} & \textbf{.602  $\pm$ .008} & \textbf{.213   $\pm$ .011} &  \textbf{.532  $\pm$ .010} & \textbf{.447  $\pm$ .010} & \textbf{ 2.02  $\pm$ .021} & \textbf{ 1.42  $\pm$ .020} \\
  \bottomrule
\end{tabular}  
\end{table*}

\subsection{Implementation Details}

We implement our proposed \Ours with PyTorch 0.4.1. For training models, we use SGD with a mini-batch of 8 patients.  
We train on 1 GPU (TITAN RTX 2080) for 50 epochs, with a learning rate of 0.001.  We randomly divide the patients in the dataset into 10 sets for each client. All the experiment results are averaged from 10-fold cross-validation, in which 7 sets are used for training, 1 set for validation, and 2 sets for testing. The validation sets are used to determine the best values of parameters in training iterations.
We run the models 20 times and report the mean and standard deviation of the metrics. We implement the propensity score prediction function $f_{\vtheta}$ and factual prediction function $g_{\vphi}$ with two-layers fully connected networks (with the size of hidden vector as 128) followed by a Sigmoid layer. We incorporate the hospital treatment assignment strategy by adding a learnable scalar variable $h_c$ to the output of the second fully connected layer. 
More details can be found at GitHub\footnote{\label{github}\url{https://github.com/yinchangchang/FED-IPTW}}.

\subsection{Individual Treatment Effect (ITE) Estimation}

\autoref{tab:result_real_ite}   displays the ITE estimation performance of state-of-the-art methods and the proposed \Ours. 
The results show that the proposed method outperforms the baselines on all the metrics in both real-world and synthetic datasets, which demonstrates the effectiveness of the proposed model. The baselines fail to consider the difference in confounding bias in various hospitals, so the ITE estimation model is likely to learn the confounding bias during local training. 

Moreover, in centralized settings, with the new variables $H$ to capture the variation
of the treatment assignment strategies across hospitals, Global outperforms Global$^{-h}$ (without modeling hospital strategy), which further demonstrates that local decorrelation can reduce confounding bias and improve federated learning algorithms' upper bound (\ie, performance in centralized settings).

\section{Discussion}

\subsection{Ablation Study}
To further demonstrate the effectiveness of the two-level decorrelations, we develop various versions of \Ours. \autoref{tab:result_real_ablation} displays the ITE estimation results in real-world and synthetic datasets respectively. The results show that \Ours outperforms IPTW$^g$ and IPTW$^l$, which demonstrates the decorrelation at both global and local levels can improve the ITE estimation performance. Moreover, \Ours outperforms \Ours$^{-h}$, which demonstrates the effectiveness of the new variable $H$.  With the new variable $H$ to capture the treatment strategy variation in each local hospital, the learned treatment prediction model becomes more accurate, which is helpful for decorrelation between $T$ and $X$, and improves the downstream ITE estimation tasks.

\subsection{Propensity Score Prediction}

The propensity score is used to adjust the confounding bias. The more accurate propensity score is helpful for the decorrelation between treatments and covariates, which can improve the downstream individual treatment effect estimation tasks. We evaluate the propensity score prediction  (\ie, treatment assignment prediction) performance as \autoref{tab:treatment_prediction} shown. The results show that with the new variable $H$ to capture treatment strategy in each hospital, the treatment prediction performance can be improved by more than 5\% in both centralized (\ie, Global$^h$) and decentralized (\ie, \Ours) settings. 
To fairly compare with \FedAvg, \Ours$^{h-train}$ hides the hospital variable $H$ (\ie, set $h_c=0$) during the test phase and only uses $h_c$ during training, \Ours$^{h-train}$ still outperforms \FedAvg, which demonstrates that capturing local treatment strategy with $H$ can reduce the bias and make the learned propensity score function more robust.

\begin{table} 
\centering
\caption{Propensity score Prediction Performance. $h_{train}$ and $h_{test}$ mean whether the variable $H$ is used during training and test phases respectively.}  
\label{tab:treatment_prediction}  
\setlength{\tabcolsep}{1pt}
\begin{tabular}{lcccccc}
\toprule 
Method   & $h_{train}$ & $h_{test}$& AUROC & AUPRC \\
\midrule
\FedAvg   & && .732 $\pm$.021 & .365 $\pm$ .013\\
\Ours$^{h-train}$& \checkmark & & .748 $\pm$.022 & .382 $\pm$ .012 \\
\Ours & \checkmark & \checkmark&  \textbf{.812$\pm$ .020} & \textbf{.425 $\pm$ .012} \\
\midrule
Global & & & .847$\pm$ .023 & .468 $\pm$ .011\\
Global$^{h-train}$& \checkmark &  & .849$\pm$ .021 & .470 $\pm$ .012\\
Global$^h$& \checkmark & \checkmark & \textbf{.889$\pm$ .022} & \textbf{.533 $\pm$ .012}\\
\bottomrule
\end{tabular}   
\end{table}

\subsubsection{Global and Local Decorrelation}

\autoref{tab:correlation} shows that FED-IPTW reduced the most correlation between $X$ and $T$, which demonstrates the effectiveness of FED-IPTW in addressing the confounding bias. 
FED-IPTW also achieved the best ITE estimation metric IF-PEHE in eICU dataset.

\begin{table}
    \centering
    \caption{Correlation and ITE estimation results in eICU dataset. $COV_l$ and $COV_g$ are local and global correlation between $X$ and $T$. 
    Note that AdaTrans, FedCI and CausalRFF do not reweight patients, so the correlation results are the same. 
    }
    \vspace{-3mm}
\setlength{\tabcolsep}{2pt}
    \begin{tabular}{l|ccc}
        \toprule
        \vspace{-1pt} 
        Methods & $COV_l$ & $COV_g$ & IF-PEHE \\
        \midrule        
AdaTrans  & .27 $\pm$ .01 & .25 $\pm$ .01 & .261   $\pm$ .013 \\
FedCI     & .27 $\pm$ .01 & .25 $\pm$ .01& .245 $\pm$ .012   \\
CausalRFF & .27 $\pm$ .01 & .25 $\pm$ .01& .231   $\pm$ .011 \\
IPTW$^{l}$ &  .06 $\pm$ .02  & .22 $\pm$ .02 &  .246 $\pm$ .012 \\  
IPTW$^{g}$ & .19 $\pm$ .02 & \textbf{.06 $\pm$ .02} &   .267 $\pm$ .011 \\ 
\midrule
\Ours & \textbf{.05 $\pm$ .02} & \textbf{.06 $\pm$ .02} &    \textbf{.213   $\pm$ .011}\\ 
         \bottomrule
    \end{tabular}
    \label{tab:correlation} 
\end{table}

%% file: 7_related_work.tex
\section{Related Work}
In this section, we briefly review the existing studies related to ITE estimation and federated learning. 
\vspace{-2mm}
\subsection{ITE Estimation}
Most existing ITE estimation studies are designed for centralized settings, including  matching-based methods \cite{crump2008nonparametric,rosenbaum1983central}, reweighting-based methods~\cite{rosenbaum1983central}, representation-based methods~\cite{johansson2016learning,shalit2017estimating}. 
In this study, we focus on the reweighting-based methods.  The reweighting-based methods attempt to re-weight samples to correct the bias in observational data. For example, IPTW~\cite{rosenbaum1983central} removes the confounding bias by assigning a weight to each individual in the population. The weights are calculated based on the propensity score.
In this work, we aim to estimate the treatment effect in decentralized settings, where the data are collected from multiple clients and are not shareable.

The existing studies most closely related to our work are federated ITE estimation models~\cite{AdaTRANS,FedCI,causalRFF}.
Federated ITE estimation aims to estimate ITE in decentralized settings while not sharing data among multiple clients.
\cite{AdaTRANS} proposes AdaTRANS to infer causal effects in the target population by utilizing multiple data sources with three levels of knowledge transfer in the inference of the outcome, treatment, and confounders. 
FedCI~\cite{FedCI} generalizes the Bayesian imputation approach~\cite{imbens2015causal} to a more generic model based on Gaussian processes (GPs) for estimating causal effects from federated observational data sources. The resulting model is decomposed into multiple components, each of which handles a distinct data source.
CausalRFF~\cite{causalRFF} leverages the Random Fourier Features for federated estimation of causal effects. The Random Fourier Features allow the objective function to be divided into multiple components to support the federated training.
\cite{han2021federated} proposes FACE to adopt a semiparametric density ratio weighting approach to estimate ITE for a flexibly specified target population of interest.

These studies achieved promising results in federated ITE estimation tasks by decomposing the objective function into multiple components and handling a distinct one in each client. However, they fail to consider the difference in the confounding coefficients (\ie, the treatment assignment strategy), which would cause the trained ITE estimation models still suffer from confounding bias during local training. We also provided theoretical proof of why the existing work cannot handle local confounding biases in various hospitals. The details can be found in \autoref{sec:two_decorrelation} in supplementary materials. 

\subsection{Federated Learning}
\noindent\textbf{Generic federated learning.}
Generic FL aims to learn a single global model, just like standard centralized learning. Since the standard algorithm FL \FedAvg~\cite{mcmahan2017communication} was introduced, many works are proposed to improve it from the aspects of local training~\cite{malinovskiy2020local,yuan2020federated,zhao2018federated,wang2020tackling,chen2021bridging}, global aggregation~\cite{yurochkin2019bayesian,wang2020federated,lin2020ensemble,chen2021fedbe,hsu2019measuring,reddi2021adaptive}, pre-training~\cite{chen2022pre}, architectures~\cite{qu2021rethinking}, etc. 

\noindent\textbf{Personalized federated learning.} 
Personalized FL~\cite{kairouz2019advances} is a recent FL paradigm that acknowledges the non-IID distributions across clients and learns personalized models for individual clients to serve on their data distributions. Many approaches have been explored to achieve it, such as federated multi-task learning~\cite{li2020federated-FMTL,smith2017federated}, representation learning~\cite{arivazhagan2019federated,collins2021exploiting,pan2024adaptive}, model interpolation~\cite{mansour2020three,chen2023federated}, fine-tuning~\cite{yu2020salvaging,chen2021bridging}.  

In an era where data privacy and accessibility are paramount, FL emerges as a promising approach, enabling collaborative AI model training across distributed healthcare datasets to improve multiple clinical tasks, such as
disease detection~\cite{yan2024fedeye}, clinical risk prediction~\cite{zhang2024unified,pan2024adaptive}, medical image segmentation~\cite{jiang2024ufps}, and ITE estimation~\cite{FedCI,causalRFF}.
In this work, we propose an interesting approach that specifically tailors the ITE estimation problem. We show that achieving an ideal generic FL model requires the facilitation of personalized FL, modeling the local conditional probability for the predictive model in \Ours weights. 

%% file: 6_conclusion.tex
\section{Conclusion} 
In this study, we investigate the treatment assignment strategy variation problem of multi-source data, which will cause inaccurate ITE estimation models.  
Especially in federated settings, such strategy variation can make the models learn the local confounding bias during local training of FL algorithms even if the global confounding bias is removed.
To address the issue, we introduce a new hospital variable $H$ to capture the local treatment assignment strategy variation and propose a new method \Ours that explicitly utilizes the new variable to decorrelate covariates and treatments both locally (within each hospital) and globally (over all the data). The resulting ITE estimation model learned via \FedAvg is thus more accurate.
We conduct extensive experiments on both synthetic and real-world datasets, and show that \Ours outperforms state-of-the-art methods.
The proposed model can provide individualized treatment decisions that can improve patient outcomes.

%% file: 5_ethic.tex



\section*{Data availability}
The datasets used in the paper are from eICU \cite{eicu} and are publicly available 
 at \url{https://eicu-crd.mit.edu/}.

\section*{Code availability}
The source code for this paper can be downloaded from the GitHub at \url{https://github.com/yinchangchang/FED-IPTW}.

%% file: appendix.tex
\appendix
\onecolumn

\renewcommand{\thefigure}{A\arabic{figure}}
\setcounter{figure}{0}
\renewcommand{\thetable}{A\arabic{table}}
\setcounter{table}{0}
\renewcommand{\theequation}{A\arabic{equation}}
\setcounter{equation}{0}
\setlength{\tabcolsep}{6pt}

\section{Supplementary Materials} 

\subsection{Assumptions}
\label{sec:assumptions}

Our estimation of ITE is based on the following important assumptions \cite{hernan2010causal}, and we further extend the assumptions in our scenario (i.e., federated setting). 
\begin{assumption}[Consistency]\label{as:consistency}
The potential outcome under treatment $T$ equals to the observed outcome if the actual treatment history is $T$.
\end{assumption}

\begin{assumption}[Positivity]\label{as:positivity}
For any patient, if the probability $P(T|X)\neq0$, then the probability of receiving treatment $0$ or $1$ is positive, i.e., $0<P(T|X)<1$, for all $X$.   
\end{assumption}

Besides these two assumptions, many existing work are based on \textit{strong ignorability} assumption:
\begin{assumption}[Strong Ignorability]\label{as:ignorability}
Given the observed historical covariates $X$, the potential outcome variables $Y$ are independent of the treatment assignment, i.e., $(y_{1},y_0)\independent T|X$
\end{assumption}

\begin{assumption}[Data Overlap]
The individuals in all sources have the same set of common covariates and there are noticeable overlap of data distribution across the client.  
\end{assumption}

Assumptions 1-3 hold in every client, so we can use our data to quantify causal effects. Assumption 4 makes the federated framework able to improve the model performance with collaboration among different clients.

\subsection{Statistics of eICU Dataset}
\label{sec:eicu_data}
\autoref{tab:statis} displays the statistics of the used eICU dataset. 

\begin{table*}[ht]
\centering
\caption{Statistics of eICU dataset.}  
\label{tab:statis} 
\begin{tabular}{ccccccc}
\toprule 
Client id & \#. Patients & \# Male/Female & Age & Treatment Rate & Mortality Rate \\
\midrule

0 & 311 & 176/135& 63.7 $\pm$ 16.7 & 33.44\% & 22.19\% \\
1 & 470 & 236/234& 64.5 $\pm$ 15.0 & 11.06\% & 12.13\% \\
2 & 246 & 119/127& 68.3 $\pm$ 15.6 & 27.24\% & 18.29\% \\
3 & 270 & 166/104& 66.5 $\pm$ 16.7 & 18.89\% &  7.41\% \\
4 & 250 & 110/140& 65.8 $\pm$ 17.4 &  3.20\% &  4.40\% \\
5 & 302 & 160/142& 64.9 $\pm$ 17.7 & 10.93\% & 15.56\% \\
6 & 393 & 205/188& 64.9 $\pm$ 14.7 & 35.88\% & 23.66\% \\
7 & 363 & 174/189& 64.6 $\pm$ 16.9 & 13.22\% & 11.02\% \\
8 & 290 & 142/148& 60.3 $\pm$ 17.0 &  6.21\% &  9.31\% \\
9 & 320 & 138/182& 64.0 $\pm$ 16.4 &  9.38\% &  8.12\% \\
10 & 321 & 148/173& 66.1 $\pm$ 15.6 & 22.43\% & 25.55\% \\
11 & 303 & 153/150& 68.4 $\pm$ 16.6 &  9.57\% &  9.24\% \\
12 & 397 & 217/180& 67.6 $\pm$ 15.3 & 14.11\% & 23.68\% \\
13 & 344 & 173/171& 74.6 $\pm$ 14.5 &  9.30\% & 13.66\% \\
14 & 211 & 124/87& 62.2 $\pm$ 17.5 & 19.43\% &  4.27\% \\
15 & 457 & 217/240& 64.8 $\pm$ 17.8 &  7.88\% & 12.69\% \\
16 & 292 & 151/141& 68.8 $\pm$ 14.8 & 14.04\% & 16.44\% \\
17 & 237 & 107/130& 71.7 $\pm$ 12.3 & 19.41\% & 12.24\% \\
18 & 228 & 115/113& 66.0 $\pm$ 15.2 & 16.67\% & 15.79\% \\
19 & 358 & 169/189& 64.8 $\pm$ 16.3 & 17.04\% & 18.16\% \\
20 & 712 & 360/352& 63.1 $\pm$ 16.7 &  0.00\% & 17.13\% \\
21 & 293 & 139/154& 64.1 $\pm$ 17.5 & 21.16\% & 15.70\% \\
22 & 332 & 160/172& 63.6 $\pm$ 16.2 & 18.37\% & 16.87\% \\
23 & 338 & 157/181& 59.1 $\pm$ 17.4 & 27.51\% & 15.68\% \\
24 & 337 & 160/177& 65.0 $\pm$ 14.3 & 12.76\% & 12.76\% \\
25 & 258 & 140/118& 62.6 $\pm$ 17.2 & 33.33\% & 12.79\% \\
26 & 325 & 158/167& 64.4 $\pm$ 17.1 & 11.38\% & 16.62\% \\

\bottomrule
\end{tabular}  
\end{table*}

\subsection{Evaluation Metrics}
\label{sec:metrics}
\textbf{Factual prediction evaluation.}
For the real-world dataset, we first evaluate the performance of factual prediction (\ie, treatment prediction and risk prediction) with area under the receiver operating characteristic curve  (AUROC) and area Under the precision-recall curve (AUPRC).

\textbf{ITE evaluation.}
To evaluate the estimated ITE, we adopt mean squared error (MSE) between the ground truth and estimated ITE as follows,
\begin{equation}
    \text{PEHE} = \frac{1}{|D|}\sum_{c=1}^{C}\sum_{i=1}^{|D_c|}((y_{c,i}(1)-y_{c,i}(0))-(\hat{y}_{c,i}(1)-\hat{y}_{c,i}(0)))^{2},
\end{equation}
which is also known as Precision in Estimation of Heterogeneous Effect (PEHE). Typically, we report the rooted PEHE ($\sqrt{\text{PEHE}}$) in our paper. We are also interested in the causal effect over the whole population to help determine whether a treatment should be assigned to population. Then we calculate the mean absolute error (MAE) between the ground truth and estimated and average treatment effect (ATE):
\begin{equation}
\begin{aligned}
    \text{MAE}=|\frac{1}{|D|}\sum_{c=1}^{C}\sum_{i=1}^{|D_c|}(y_{c,i}(1) -y_{c,i}(0))  - \frac{1}{|D|}\sum_{c=1}^{C}\sum_{i=1}^{|D_c|}(\hat{y}_{c,i}(1)-\hat{y}_{c,i}(0))|  
\end{aligned}
\end{equation}

As the true treatment effects are not available in real-world data, we use the influence function-based precision of estimating heterogeneous effects (IF-PEHE) for model evaluation. We calculate IF-PEHE as,

\begin{itemize}
    \item Step 1: Train two XGBoost classifiers for potential outcome prediction denoted by $\mu_0$ and $\mu_1$, where $\mu_a = P(y_a = 1|X = x)$ using the training set.
    Then calculate the plug-in estimation $e = \mu_1 - \mu_0$. 
    Train a XGBoost classifier propensity score function (i.e., the probability of receiving treatment) $\pi_t = P(T = 1|X = x)$.
    \item Step 2:  Given the estimated treatment effect $\hat{e}(x_i)$ on the test set, calculate the IF-PEHE with the influence function $\hat{l}$ as,
\end{itemize}
\begin{align}
    & \text{IF-PEHE} = \sum_{i} [(\hat{e}(x_i) - e(x_i))^2 + \hat{l}(x_i)], \\
    & \quad \text{where } \quad \hat{l}(x) = (1 -B) \hat{e}^2(x) + By(e(x) - \hat{e}(x)) \nonumber   - W(e(x) - \hat{e}(x))^2 + \hat{e}^2(x), \nonumber \\
    & \quad \qquad \qquad W = (t - \pi(x)), B = 2t(t - \pi(x))Z^{-1},  Z=\pi(x)(1-\pi(x)) \nonumber
\end{align}

\subsection{Global and Local Correlation}
\label{sec:two_decorrelation}
We define the global correlation $COV(X, T)$ and local correlation $COV(X_c, T_c)$ as follows:

\begin{align}
& COV(X,T) = \sum_{c=1}^C \sum_{i=1}^{|D_c|} w_{c,i} (x_{c,i} - \overline{X}) (t_{c,i} - \overline{T}),     \label{eq:global_correlation_s} \\
& \hspace{5pt} \text{where } \overline{X} = \frac{1}{w^s}\sum_{c=1}^C \sum_{i=1}^{|D_c|} w_{c,i} x_{c,i}, \quad  
        \overline{T} = \frac{1}{w^s}\sum_{c=1}^C \sum_{i=1}^{|D_c|} w_{c,i} t_{c,i}, \nonumber 
\end{align}
\begin{align}
& \hspace{30pt}  w^s = \sum_{c=1}^C w^s_c, \quad w^s_c = \sum_{i=1}^{|D_c|} w_{c,i}. \nonumber\\ %
& COV(X_c,T_c) = \sum_{i=1}^{|D_c|} w_{c,i} (x_{c,i} - \overline{X_c}) (t_{c,i} - \overline{T_c}), \label{eq:local_correlation_s} \\
& \hspace{5pt} \text{where } \overline{X_c} =  \frac{1}{w^s_c}\sum_{i=1}^{|D_c|} w_{c,i} x_{c,i}, \quad 
        \overline{T_c} = \frac{1}{w^s_c}\sum_{i=1}^{|D_c|} w_{c,i} t_{c,i}. \nonumber 
\end{align}
$\overline{X_c}$ and $\overline{T_c}$ are the average covariates and treatments in client $c$. 
$\overline{X}$ and $\overline{T}$ are the average covariates and treatments among all the clients. 

The patient-specific weight $w_{c,i}$ is computed as:
\begin{equation}
\hspace{-2mm} w_{c,i} = \frac{E[T_c] * I[t_i=1]}{ P(T=1|x_{c,i},h_c)} + \frac{(1 - E[T_c]) * I[t_i=0]}{P(T=0|x_{c,i},h_c) }
\end{equation}

Based on the observed value of $t_{c,i}$, the patient-specific weight can be re-written as:

\begin{equation}
w^{(1)}_{c,i} = \frac{E[T_c]}{P(T=1|x_{c,i}, h_c)}, w^{(0)}_{c,i} = \frac{1 - E[T_c]}{P(T=0|x_{c,i},h_c)},
\end{equation} 
where $w^{(1)}_{c,i}, w^{(0)}_{c,i}$ are weights for patients in treated and control groups.

We first show that the weight sum $w^s_c$ in each client is equal to $|D_c|$ and $w^s$ is equal to $|D|$:

\begin{align}
    & w^s_c = \sum_{i=1}^{|D_c|} w_{c,i} \nonumber  =  |D_c| \frac{1}{|D_c|}\sum_{i=1}^{|D_c|} w_{c,i} \nonumber \\  
    \nonumber & = |D_c| \int_{x} [\frac{E[T_c]}{P(T=1|x, h_c)} P(T=1|x, h_c)  + \frac{1 - E[T_c]}{P(T=0|x,h_c)} P(T=0|x, h_c) ]  dF_{X_c}(x) \\  
    \nonumber & = |D_c|\int_x [E[T_c] + (1 - E[T_c])] dF_{X_c}(x)  = |D_c| \int_x  dF_{X_c}(x)   = |D_c|,
\end{align}
\hspace{-2mm} where $F_{X_c}(x)$ is the probability density ratio of $X$ in client $c$.

\begin{align}
    w^s = \sum_{c=1}^C w^s_c  
    = \sum_{c=1}^C |D_c| 
    = |D| \qquad \qquad \qquad \quad \nonumber
\end{align}

We found that the patient-specific weight $w_{c,i}$ in \Ours does not change the average of covariates (\ie, $\overline{X}_c$ and $\overline{X}$) and treatments (\ie, $\overline{T}_c$ and $\overline{T}$).

\begin{align}
    \nonumber & \overline{X_c} = \frac{1}{w^s_c}\sum_{i=1}^{|D_c|} w_{c,i} x_{c,i}  = \frac{1}{|D_c|}\sum_{i=1}^{|D_c|} w_{c,i} x_{c,i}  \\
    & \qquad = \int_{x} [\frac{E[T_c]}{P(T=1|x, h_c)} P(T=1|x, h_c)  + \frac{1 - E[T_c]}{P(T=0|x,h_c)} P(T=0|x, h_c) ] x  ~dF_{X_c}(x)\\
    \nonumber & \qquad= \int_{x} [E[T_c] + (1 - E[T_c])] x  ~dF_{X_c}(x)  = \int_{x} x ~dF_{X_c}(x)  = E[X_c]
\end{align}

\begin{align}
    \nonumber & \overline{T_c} = \frac{1}{w^s_c}\sum_{i=1}^{|D_c|} w_{c,i} t_{c,i}  = \frac{1}{|D_c|}\sum_{i=1}^{|D_c|} w_{c,i} t_{c,i} 
    =  \int_{x} \frac{E[T_c]}{P(T=1|x, h_c)} P(T=1|x, h_c) dF_{X_c}(x)  
    \\ & =\int_{x}  E[T_c]  dF_{X_c}(x)= E[T_c]
\end{align}

\begin{align}
    & \overline{X} = \frac{1}{w^s}\sum_{c=1}^C \sum_{i=1}^{|D_c|} w_{c,i} x_{c,i}  = \frac{1}{w^s}\sum_{c=1}^C w^s_c E[X_c]  = \frac{1}{|D|}\sum_{c=1}^C |D_c| E[X_c]\nonumber   = E[X] \nonumber \\
    & \overline{T} = \frac{1}{w^s}\sum_{c=1}^C \sum_{i=1}^{|D_c|} w_{c,i} t_{c,i}  = \frac{1}{w^s}\sum_{c=1}^C w^s_c E[T_c] = \frac{1}{|D|}\sum_{c=1}^C |D_c| E[T_c]  = E[T] \nonumber
\end{align}

With the above definition of weights and weighted average of treatments and covariates, we proceed to present proofs for both local decorrelation and global decorrelation.
\subsubsection{Local Decorrelation}
We demonstrate that the local correlation in \autoref{eq:local_correlation_result} converges to 0 when utilizing the patient-specific weight in \Ours:

\begin{align}    
& \hspace{-6mm} COV(X_c, T_c) = \sum_{i=1}^{|D_c|} w_{c,i} (x_{c,i} - \overline{X_c}) (t_{c,i} - \overline{T_c}) = |D_c| \frac{1}{|D_c|} \sum_{i=1}^{|D_c|} w_{c,i} (x_{c,i} - \overline{X_c}) (t_{c,i} - \overline{T_c}) \nonumber\\ 
& \hspace{-4mm} = |D_c| \int_x [\frac{E[T_c]}{P(T=1|x,h_c)} (x - \overline{X}_c) (1- \overline{T}_c) P(T=1|x,h_c)  \nonumber \\ & \qquad + \frac{1 - E[T_c]}{P(T=0|x,h_c)} (x-\overline{X}_c) (0- \overline{T}_c) P(T=0|x,h_c) ] dF_{X_c}(x) \nonumber\\
& \hspace{-4mm} = |D_c| \int_x [E[T_c] (x - \overline{X}_c) (1- \overline{T}_c)  + (1 - E[T_c]) (x-\overline{X}_c) (0- \overline{T}_c)  ] dF_{X_c}(x) \nonumber \\
& \hspace{-4mm} = |D_c| \int_x (x - \overline{X}_c) [E[T_c]  (1- \overline{T}_c)  + (1 - E[T_c])   (0- \overline{T}_c)  ] dF_{X_c}(x) \nonumber\\ 
& \hspace{-4mm} = |D_c| \int_x (x - \overline{X}_c) [E[T_c] - \overline{T}_c  ] dF_{X_c}(x) \nonumber\\ 
& \hspace{-4mm} = |D_c| \int_x (x - \overline{X}_c) [E[T_c] - E[T_c]  ] dF_{X_c}(x) \nonumber\\ 
& \hspace{-4mm} = \sum_{i=1}^{|D_c|} (x_{c,i} - \overline{X}_c) [E[T_c] - E[T_c]  ]  = 0 \label{eq:app:local_decorrelation}
\end{align}

\subsubsection{Global decorrelation}

Taking into account solely the patient-specific weight $w_{c,i}$, the calculation for global correlation is as follows:

\begin{align}
& \hspace{-2mm} COV(X,T) = \sum_{c=1}^C \sum_{i=1}^{|D_c|} w_{c,i} (x_{c,i} - \overline{X}) (t_{c,i} - \overline{T}) \nonumber
\end{align}
 
\vspace{-2mm}

We compute each item in $COV(X,T)$ as $COV_c(X,T)$: 

\begin{align}
& COV_c(X,T) = \sum_{i=1}^{|D_c|} w_{c,i} (x_{c,i} - \overline{X}) (t_{c,i} - \overline{T}) \nonumber\\
& = |D_c| \frac{1}{|D_c|} \sum_{i=1}^{|D_c|} w_{c,i} (x_{c,i} - \overline{X}) (t_{c,i} - \overline{T}) \nonumber \\
&  = |D_c| \int_x [\frac{E[T_c]}{P(T=1|x,h_c)} (x - \overline{X}) (1- \overline{T}) P(T=1|x,h_c) \nonumber \\
& \qquad + \frac{1 - E[T_c]}{P(T=0|x,h_c)} (x-\overline{X}) (0- \overline{T}) P(T=0|x,h_c) ] dF_{X_c}(x) \nonumber \\
&  = |D_c| \int_x [E[T_c] (x - \overline{X}) (1- \overline{T}) + (1 - E[T_c]) (x-\overline{X}) (0- \overline{T})  ] dF_{X_c}(x) \nonumber\\
&  = |D_c| \int_x (x - \overline{X}) [E[T_c]  (1- \overline{T})  + (1 - E[T_c])   (0- \overline{T})  ] dF_{X_c}(x) \nonumber\\ 
&  = |D_c| \int_x (x - \overline{X}) [E[T_c] - \overline{T}  ] dF_{X_c}(x) \nonumber\\ 
&  = \sum_{i=1}^{|D_c|} (x_{c,i} - \overline{X}) (E[T_c] - \overline{T}  ) \nonumber \\ 
&  = |D_c|  (E[X_c] - \overline{X}) (E[T_c] - \overline{T}  ) \nonumber \qquad\qquad\qquad\qquad\\ 
&  =  |D_c| (E[X_c] - E[X]) (E[T_c] - E[T] )
\end{align}

\vspace{2mm}

$COV(X,T)$ can be computed by summing $COV_c(X,T)$ and we have the \autoref{eq:global_correlation_wo_wc}:
\begin{align}
& \hspace{-2mm} COV(X,T) = \sum_{c=1}^{C}  COV_c(X,T)   = \sum_{c=1}^{C} |D_c|  (E[X_c] - E[X]) (E[T_c] - E[T] ) \nonumber 
\end{align}

We compute a hospital-specific weight $w_c$ as follows:


\begin{align}
\label{eq:hospital_weight_s}
w_c = \frac{P(T' = E[T_c])}{P(T'=E[T_c]|X'=E[X_c])}
\end{align}

\hspace{2mm}
 
With the hospital-specific weight $w_c$, the global correlation in \autoref{eq:global_correlation_0} is removed.

\begin{align}
& \hspace{-1mm} COV'(X,T) = \sum_{c=1}^C w_c \sum_{i=1}^{|D_c|} w_{c,i} (x_{c,i} - \overline{X}) (t_{c,i} - \overline{T}) \nonumber \\ 
& =  \sum_{c=1}^C w_c |D_c| (E[X_c] - E[X]) (E[T_c] - E[T]) \nonumber \\
& =  \int_x \int_t w_c |D_c| (x - E[X]) (t - E[T]) dF_{T'|X'}(t)  dF_{X'}(x) \nonumber  \\
& = \int_x \int_t \frac{P(T' = t)}{P(T'=t|X'=x)} |D_c| (x - E[X]) (t - E[T]) * P(T'=t|X'=x)P(X'=x) dt dx \nonumber \\ 
& = \int_x \int_t P(T' = t) |D_c|  (x - E[X]) (t - E[T])  * P(X'=x) dt dx \nonumber \\
& = \int_t\int_x  |D_c|  (x - E[X])  P(X'=x) dx * P(T' = t) (t - E[T]) dt\nonumber \\
& = \int_t (E[X] - E[X]) P(T'=t)dt \nonumber \\
& = 0  \label{eq:app:global_decorrelation}
\end{align}

\subsection{Hospital-Specific Weight Computation}
\label{sec:hospital_weight}
According to the central limit theorem, for independent and identically distributed random variables, the sampling distribution of the standardized sample mean tends towards the Gaussian distribution.

We model the $P(T')$ and $P(T'|X')$ in \autoref{eq:hospital_weight} with Gaussian distributions. $T'$ and $X'$ denote the averages of treatments and covariates respectively for hospitals.

We assume $T' \sim N (\mu_0, \sigma_0^2)$. The parameters are learned as follows:
\begin{align}
    & \mu_0 = E[T],  \sigma_0^2 = \frac{1}{|D|} \sum_{c=1}^C |D_c| (E[T_c] - E[T])^2
\end{align}

Given a client $c$, we compute the treatment average probability as follows:\
\begin{equation}
\label{eq:prob_T_average}
    P(T'=E[T_c]) = \frac{1}{\sigma_0 \sqrt{2\pi}}e^{-\frac{1}{2}(\frac{E[T_c] - \mu_0}{\sigma_0})^2}
\end{equation}

We use the Gaussian process regression (GPR) to model the posterior distribution $P(T'|X')$.
We use a GPR model to fit the dataset $\{(E[X_c], E[T_c])| c = 1, 2, ..., C\}$.
Given each client $c$, GPR can generate the parameters $\mu_c$ and $\sigma_c$. 
Then we compute the treatment average probability given the covariate average as follows:
\begin{equation}
\label{eq:post_T_X_average}
    P(T'=E[T_c]|X'=E[X_c]) = \frac{1}{\sigma_c \sqrt{2\pi}}e^{-\frac{1}{2}(\frac{E[T_c] - \mu_c}{\sigma_c})^2}
\end{equation}

With the two probabilities in \autoref{eq:prob_T_average} and \autoref{eq:post_T_X_average}, we can compute the hospital-specific weight in \autoref{eq:hospital_weight}.

%% file: main.bbl
\begin{thebibliography}{}

\bibitem[Arivazhagan et~al., 2019]{arivazhagan2019federated}
Arivazhagan, M.~G., Aggarwal, V., Singh, A.~K., and Choudhary, S. (2019).
\newblock Federated learning with personalization layers.
\newblock {\em arXiv preprint arXiv:1912.00818}.

\bibitem[Cadene et~al., 2019]{cadene2019rubi}
Cadene, R., Dancette, C., Cord, M., Parikh, D., et~al. (2019).
\newblock Rubi: Reducing unimodal biases for visual question answering.
\newblock {\em Advances in neural information processing systems}, 32.

\bibitem[Chen and Chao, 2021]{chen2021fedbe}
Chen, H.-Y. and Chao, W.-L. (2021).
\newblock Fedbe: Making bayesian model ensemble applicable to federated learning.
\newblock In {\em ICLR}.

\bibitem[Chen and Chao, 2022]{chen2021bridging}
Chen, H.-Y. and Chao, W.-L. (2022).
\newblock On bridging generic and personalized federated learning for image classification.
\newblock In {\em ICLR}.

\bibitem[Chen et~al., 2023a]{chen2022pre}
Chen, H.-Y., Tu, C.-H., Li, Z., Shen, H.-W., and Chao, W.-L. (2023a).
\newblock On the importance and applicability of pre-training for federated learning.
\newblock In {\em ICLR}.

\bibitem[Chen et~al., 2023b]{chen2023federated}
Chen, H.-Y., Zhong, J., Zhang, M., Jia, X., Qi, H., Gong, B., Chao, W.-L., and Zhang, L. (2023b).
\newblock Federated learning of shareable bases for personalization-friendly image classification.
\newblock {\em arXiv preprint arXiv:2304.07882}.

\bibitem[Collins et~al., 2021]{collins2021exploiting}
Collins, L., Hassani, H., Mokhtari, A., and Shakkottai, S. (2021).
\newblock Exploiting shared representations for personalized federated learning.
\newblock In {\em International Conference on Machine Learning}, pages 2089--2099. PMLR.

\bibitem[Crump et~al., 2008]{crump2008nonparametric}
Crump, R.~K., Hotz, V.~J., Imbens, G.~W., and Mitnik, O.~A. (2008).
\newblock Nonparametric tests for treatment effect heterogeneity.
\newblock {\em The Review of Economics and Statistics}, 90(3):389--405.

\bibitem[Han et~al., 2021]{han2021federated}
Han, L., Hou, J., Cho, K., Duan, R., and Cai, T. (2021).
\newblock Federated adaptive causal estimation (face) of target treatment effects.
\newblock {\em arXiv preprint arXiv:2112.09313}.

\bibitem[Hernan and Robins, 2010]{hernan2010causal}
Hernan, M.~A. and Robins, J.~M. (2010).
\newblock Causal inference.

\bibitem[Hsu et~al., 2019]{hsu2019measuring}
Hsu, T.-M.~H., Qi, H., and Brown, M. (2019).
\newblock Measuring the effects of non-identical data distribution for federated visual classification.
\newblock {\em arXiv preprint arXiv:1909.06335}.

\bibitem[Imbens and Rubin, 2015]{imbens2015causal}
Imbens, G.~W. and Rubin, D.~B. (2015).
\newblock {\em Causal inference in statistics, social, and biomedical sciences}.
\newblock Cambridge University Press.

\bibitem[Jiang et~al., 2024]{jiang2024ufps}
Jiang, L., Ma, L.~Y., Zeng, T.~Y., and Ying, S.~H. (2024).
\newblock Ufps: A unified framework for partially annotated federated segmentation in heterogeneous data distribution.
\newblock {\em Patterns}, 5(2).

\bibitem[Johansson et~al., 2016]{johansson2016learning}
Johansson, F., Shalit, U., and Sontag, D. (2016).
\newblock Learning representations for counterfactual inference.
\newblock In {\em ICML'16}, pages 3020--3029.

\bibitem[Kairouz et~al., 2019]{kairouz2019advances}
Kairouz, P., McMahan, H.~B., Avent, B., Bellet, A., Bennis, M., Bhagoji, A.~N., Bonawitz, K., Charles, Z., Cormode, G., Cummings, R., et~al. (2019).
\newblock Advances and open problems in federated learning.
\newblock {\em arXiv preprint arXiv:1912.04977}.

\bibitem[Kulkarni et~al., 2020]{Kulkarni2020SurveyOP}
Kulkarni, V., Kulkarni, M., and Pant, A. (2020).
\newblock Survey of personalization techniques for federated learning.
\newblock {\em 2020 Fourth World Conference on Smart Trends in Systems, Security and Sustainability (WorldS4)}, pages 794--797.

\bibitem[Li et~al., 2020a]{li2020federated-FMTL}
Li, T., Hu, S., Beirami, A., and Smith, V. (2020a).
\newblock Ditto: Fair and robust federated learning through.
\newblock {\em arXiv preprint arXiv:2012.04221}.

\bibitem[Li et~al., 2020b]{li2020federated-survey}
Li, T., Sahu, A.~K., Talwalkar, A., and Smith, V. (2020b).
\newblock Federated learning: Challenges, methods, and future directions.
\newblock {\em IEEE Signal Processing Magazine}, 37(3):50--60.

\bibitem[Lin et~al., 2020]{lin2020ensemble}
Lin, T., Kong, L., Stich, S.~U., and Jaggi, M. (2020).
\newblock Ensemble distillation for robust model fusion in federated learning.
\newblock In {\em NeurIPS}.

\bibitem[Liu et~al., 2020]{liu2020estimating}
Liu, R., Yin, C., and Zhang, P. (2020).
\newblock Estimating individual treatment effects with time-varying confounders.
\newblock In {\em 2020 IEEE International Conference on Data Mining (ICDM)}, pages 382--391. IEEE.

\bibitem[Malinovskiy et~al., 2020]{malinovskiy2020local}
Malinovskiy, G., Kovalev, D., Gasanov, E., Condat, L., and Richtarik, P. (2020).
\newblock From local sgd to local fixed-point methods for federated learning.
\newblock In {\em ICML}.

\bibitem[Mansour et~al., 2020]{mansour2020three}
Mansour, Y., Mohri, M., Ro, J., and Suresh, A.~T. (2020).
\newblock Three approaches for personalization with applications to federated learning.
\newblock {\em arXiv preprint arXiv:2002.10619}.

\bibitem[McMahan et~al., 2017a]{FedAvg}
McMahan, B., Moore, E., Ramage, D., Hampson, S., and y~Arcas, B.~A. (2017a).
\newblock Communication-efficient learning of deep networks from decentralized data.
\newblock In {\em Artificial intelligence and statistics}, pages 1273--1282. PMLR.

\bibitem[McMahan et~al., 2017b]{mcmahan2017communication}
McMahan, H.~B., Moore, E., Ramage, D., Hampson, S., et~al. (2017b).
\newblock Communication-efficient learning of deep networks from decentralized data.
\newblock In {\em AISTATS}.

\bibitem[Pan et~al., 2024]{pan2024adaptive}
Pan, W., Xu, Z., Rajendran, S., and Wang, F. (2024).
\newblock An adaptive federated learning framework for clinical risk prediction with electronic health records from multiple hospitals.
\newblock {\em Patterns}, 5(1).

\bibitem[Peine et~al., 2021]{peine2021development}
Peine, A., Hallawa, A., Bickenbach, J., Dartmann, G., Fazlic, L.~B., Schmeink, A., Ascheid, G., Thiemermann, C., Schuppert, A., Kindle, R., et~al. (2021).
\newblock Development and validation of a reinforcement learning algorithm to dynamically optimize mechanical ventilation in critical care.
\newblock {\em NPJ digital medicine}, 4(1):32.

\bibitem[Pollard et~al., 2018]{eicu}
Pollard, T.~J., Johnson, A.~E., Raffa, J.~D., Celi, L.~A., Mark, R.~G., and Badawi, O. (2018).
\newblock The eicu collaborative research database, a freely available multi-center database for critical care research.
\newblock {\em Scientific data}, 5(1):1--13.

\bibitem[Qian et~al., 2021]{qian2021synctwin}
Qian, Z., Zhang, Y., Bica, I., Wood, A., and van~der Schaar, M. (2021).
\newblock Synctwin: Treatment effect estimation with longitudinal outcomes.
\newblock {\em Advances in Neural Information Processing Systems}, 34:3178--3190.

\bibitem[Qu et~al., 2021]{qu2021rethinking}
Qu, L., Zhou, Y., Liang, P.~P., Xia, Y., Wang, F., Fei-Fei, L., Adeli, E., and Rubin, D. (2021).
\newblock Rethinking architecture design for tackling data heterogeneity in federated learning.
\newblock {\em arXiv preprint arXiv:2106.06047}.

\bibitem[Reddi et~al., 2021]{reddi2021adaptive}
Reddi, S., Charles, Z., Zaheer, M., Garrett, Z., Rush, K., Kone{\v{c}}n{\`y}, J., Kumar, S., and McMahan, H.~B. (2021).
\newblock Adaptive federated optimization.
\newblock In {\em ICLR}.

\bibitem[Rosenbaum and Rubin, 1983]{rosenbaum1983central}
Rosenbaum, P.~R. and Rubin, D.~B. (1983).
\newblock The central role of the propensity score in observational studies for causal effects.
\newblock {\em Biometrika}, 70(1):41--55.

\bibitem[Shalit et~al., 2017]{shalit2017estimating}
Shalit, U., Johansson, F.~D., and Sontag, D. (2017).
\newblock Estimating individual treatment effect: generalization bounds and algorithms.
\newblock In {\em International conference on machine learning}, pages 3076--3085. PMLR.

\bibitem[Smith et~al., 2017]{smith2017federated}
Smith, V., Chiang, C.-K., Sanjabi, M., and Talwalkar, A.~S. (2017).
\newblock Federated multi-task learning.
\newblock In {\em NeurIPS}.

\bibitem[Vo et~al., 2022a]{causalRFF}
Vo, T.~V., Bhattacharyya, A., Lee, Y., and Leong, T.-Y. (2022a).
\newblock An adaptive kernel approach to federated learning of heterogeneous causal effects.
\newblock {\em Advances in Neural Information Processing Systems}, 35:24459--24473.

\bibitem[Vo et~al., 2022b]{FedCI}
Vo, T.~V., Lee, Y., Hoang, T.~N., and Leong, T.-Y. (2022b).
\newblock Bayesian federated estimation of causal effects from observational data.
\newblock In {\em Uncertainty in Artificial Intelligence}, pages 2024--2034. PMLR.

\bibitem[Vo et~al., 2022c]{AdaTRANS}
Vo, T.~V., Wei, P., Hoang, T.~N., and Leong, T.~Y. (2022c).
\newblock Adaptive multi-source causal inference from observational data.
\newblock In {\em Proceedings of the 31st ACM International Conference on Information \& Knowledge Management}, pages 1975--1985.

\bibitem[Wang et~al., 2020a]{wang2020federated}
Wang, H., Yurochkin, M., Sun, Y., Papailiopoulos, D., and Khazaeni, Y. (2020a).
\newblock Federated learning with matched averaging.
\newblock In {\em ICLR}.

\bibitem[Wang et~al., 2020b]{wang2020tackling}
Wang, J., Liu, Q., Liang, H., Joshi, G., and Poor, H.~V. (2020b).
\newblock Tackling the objective inconsistency problem in heterogeneous federated optimization.
\newblock In {\em NeurIPS}.

\bibitem[Yan et~al., 2024]{yan2024fedeye}
Yan, B., Cao, D., Jiang, X., Chen, Y., Dai, W., Dong, F., Huang, W., Zhang, T., Gao, C., Chen, Q., et~al. (2024).
\newblock Fedeye: A scalable and flexible end-to-end federated learning platform for ophthalmology.
\newblock {\em Patterns}, 5(2).

\bibitem[Yao et~al., 2018]{yao2018representation}
Yao, L., Li, S., Li, Y., Huai, M., Gao, J., and Zhang, A. (2018).
\newblock Representation learning for treatment effect estimation from observational data.
\newblock In {\em NeurIPS'18}, pages 2633--2643.

\bibitem[Yin et~al., 2022]{yin2022deconfounding}
Yin, C., Liu, R., Caterino, J., and Zhang, P. (2022).
\newblock Deconfounding actor-critic network with policy adaptation for dynamic treatment regimes.
\newblock In {\em Proceedings of the 28th ACM SIGKDD Conference on Knowledge Discovery and Data Mining}, pages 2316--2326.

\bibitem[Yu et~al., 2020]{yu2020salvaging}
Yu, T., Bagdasaryan, E., and Shmatikov, V. (2020).
\newblock Salvaging federated learning by local adaptation.
\newblock {\em arXiv preprint arXiv:2002.04758}.

\bibitem[Yuan and Ma, 2020]{yuan2020federated}
Yuan, H. and Ma, T. (2020).
\newblock Federated accelerated stochastic gradient descent.
\newblock In {\em NeurIPS}.

\bibitem[Yurochkin et~al., 2019]{yurochkin2019bayesian}
Yurochkin, M., Agarwal, M., Ghosh, S., Greenewald, K., Hoang, T.~N., and Khazaeni, Y. (2019).
\newblock Bayesian nonparametric federated learning of neural networks.
\newblock In {\em ICML}.

\bibitem[Zhang et~al., 2024]{zhang2024unified}
Zhang, F., Shuai, Z., Kuang, K., Wu, F., Zhuang, Y., and Xiao, J. (2024).
\newblock Unified fair federated learning for digital healthcare.
\newblock {\em Patterns}, 5(1).

\bibitem[Zhao et~al., 2018]{zhao2018federated}
Zhao, Y., Li, M., Lai, L., Suda, N., Civin, D., and Chandra, V. (2018).
\newblock Federated learning with non-iid data.
\newblock {\em arXiv preprint arXiv:1806.00582}.

\end{thebibliography}
